\documentclass[conference]{IEEEtran}
\IEEEoverridecommandlockouts

\usepackage{cite}
\usepackage{amsmath,amssymb,amsfonts}
\usepackage{graphicx}
\graphicspath{ {./images/} }
\usepackage{textcomp}
\usepackage{xcolor}
\usepackage{booktabs, longtable, multirow}
\usepackage{adjustbox}
\usepackage{float}
\usepackage{dblfloatfix}
\usepackage{algorithm}
\usepackage{algpseudocode}

\def\BibTeX{{\rm B\kern-.05em{\sc i\kern-.025em b}\kern-.08em
    T\kern-.1667em\lower.7ex\hbox{E}\kern-.125emX}}
\begin{document}

\title{Evaluating Multi-Global Server Architecture for Federated Learning
\\
}


\author{\IEEEauthorblockN{Asfia Kawnine\IEEEauthorrefmark{1}, Hung Cao\IEEEauthorrefmark{1}, Atah Nuh Mih\IEEEauthorrefmark{1}, Monica Wachowicz\IEEEauthorrefmark{1}\IEEEauthorrefmark{3}}

\IEEEauthorblockA{\IEEEauthorrefmark{1} \textit{Analytics Everywhere Lab, University of New Brunswick, Canada} \\
\IEEEauthorrefmark{3} \textit{RMIT University, Australia} \\\\}
}

\maketitle

\begin{abstract}

Federated learning (FL) with a single global server framework is currently a popular approach for training machine learning models on decentralized environment, such as mobile devices and edge devices. However, the centralized server architecture poses a risk as any challenge on the central/global server would result in the failure of the entire system. To minimize this risk, we propose a novel federated learning framework that leverages the deployment of multiple global servers.
We posit that implementing multiple global servers in federated learning can enhance efficiency by capitalizing on local collaborations and aggregating knowledge, and the error tolerance in regard to communication failure in the single server framework would be handled.
We therefore propose a novel framework that leverages the deployment of multiple global servers.
We conducted a series of experiments using a dataset containing the event history of electric vehicle (EV) charging at numerous stations. We deployed a federated learning setup with multiple global servers and client servers, where each client-server strategically represented a different region and a global server was responsible for aggregating local updates from those devices.
Our preliminary results of the global models demonstrate that the difference in performance attributed to multiple servers is less than 1\%. While the hypothesis of enhanced model efficiency was not as expected, the rule for handling communication challenges added to the algorithm could resolve the error tolerance issue. Future research can focus on identifying specific uses for the deployment of multiple global servers.

\end{abstract}

\begin{IEEEkeywords}
Federated Learning, Edge AI, Multiple global servers, EV energy consumption
\end{IEEEkeywords}

\section{Introduction} \label{intro}
Federated learning (FL) is a distributed learning approach that allows machine learning models on decentralized devices while preserving data privacy\cite{mcmahan2017communication}.
In recent years, there has been significant research and development focused on addressing challenges and improving various aspects of FL.
These advancements aim to enhance the performance, efficiency, and scalability of decentralized machine learning paradigms while preserving data privacy.
The works collectively contribute to advancing the field of FL by addressing various challenges and improving different aspects of these decentralized learning paradigms.
However, the FL approaches in these works implement one central parameter server either at cloud or edge and multiple client devices. 
This reliance on a single central server for model aggregation limit the fault tolerance of the approaches as any breakdown in communication with the central server can result in the collapse of the whole architecture. 


To address these challenges, we conduct a study, which hypothesizes that implementing a multi-global server architecture in federated learning can improve training efficiency and model accuracy.
We introduce a multi-global server architecture in which multiple servers collaborate in the aggregation process, with the purpose of distributing the computational load and facilitating parallel model updates.
This approach also reduces the risks of any communication disruption in the FL architecture by providing an alternative server if any of the global servers is not available.
Our multi-global server architecture has the potential to improve model accuracy by leveraging the diversity of local models and knowledge across different servers. This can be achieved by combining the knowledge of multiple global servers through collaborative model aggregation and parameter exchange.

The primary objectives of this research are (1) to design and implement a multi-global server architecture for federated learning; (2) to evaluate the performance of the multi-global server approach in terms of training efficiency, and model accuracy; and (3) to ensure reliable communication between the multi-global server and client devices if any challenge occurs. By investigating these research objectives, we aim to provide insights into the potential benefits and trade-offs of employing a multi-global server architecture in federated learning and contribute to the advancement of scalable and efficient distributed machine learning systems.

The paper is structured as follows: Section \ref{related_work} reviews the federated learning approach and relevant related works; Section \ref{preli} includes a brief discussion on methods of FL; Section \ref{prop_meth} discusses our proposed method; Section \ref{exp} describes our experiment, including dataset and system architecture. We discuss our results and findings in Section \ref{result}; and provide a brief summary and possible future direction in Section \ref{conc}.
\section{Related Work} \label{related_work}
\subsection{Federated Learning (FL)}
Federated learning is a decentralized privacy preserving approach introduced by Google \cite{mcmahan2017communication}.
The approach can be defined as the combination of distributed learning and ensemble learning, with local learning schemes that are referred to as parallel learning schemes \cite{niknam2020federated}.
FL became popular in various applications because of its wireless communication approach.  
It was first introduced to handle large amount of data, maintain privacy, and tackle non independent and identically distributed (non iid) data from multiple sources.
Over time, FL found widespread use in edge computing, blockchain, deep networks, industrial application etc.
In addition, federated learning is integrated into egde-driven IoT systems, which aims to distribute edge learning tasks by locating the IoT devices with appropriate data \cite{xiang4330541edge}.
The research proposed a new search algorithm based on federated learning and based on the outcome it associates the new context with the IoT knowledge graph.
Another study shows that federated learning can successfully predict how much energy is consumed by households as well as how much solar production is possible \cite{bharadwaj2023energy}. In the study, FL is implemented in low-power consumed and low-memory embedded devices within a single network.
\subsection{Aggregate Strategies of FL}
Since its introduction, much progress has been achieved in different sectors of the federated learning algorithm to achieve better optimization, performance, efficiency, scalability of devices, etc.
One variant of FL that addresses the limitations of traditional FL is hierarchical federated learning (HFL) \cite{liu2020client}.
HFL introduces a hierarchical structure where intermediate aggregators, such as edge devices or clusters of devices are included in the learning process.
This approach provides additional benefits, such as reducing communication overhead, improving scalability, and allowing for more fine-grained control over the learning process.
Heterogeneous federated learning was introduced as another federated learning approach involving different feature spaces \cite{gao2022survey} by integrating different data spaces, model heterogeneity, and other features.

Apart from the different variants FL also has multiple model aggregation methods for combining the local models and also it uses various machine learning models to train the local model.
Federated averaging (FedAvg) is the most common and widely used aggregation method among them \cite{jhunjhunwala2023fedexp}.
It is a communication efficient algorithm, that aggregates the client updated using a weighted average to produce the global model.
Federated Averaging with Server Momentum (FedAvgM) is an extension of FedAvg that uses a momentum term on the server to track the historical gradients of the model and smooth out the updates \cite{reddi2020adaptive}. 
Federated AdaGrad (FedAdaGrad), Federated Yogi (FedYogi) and Federated Adam (FedAdam) are few more adaptive optimizer that are used in different heterogeneous devices.

Depending on the size, distribution, and other criteria of a dataset the models may perform differently.
The dataset we used for this study to complete the experiment does not need extra optimization, for which we selected the most used aggregation model fedAvg.
For training the local models different algorithms can be used, such  as linear models, decision trees, neural networks etc \cite{li2021survey}.
The models have been reinvented and developed according to the federated learning, so that certain weights can be extracted and sent to the global model.

\subsection{Edge and FL}
Federated learning has multiple applications and different distributed computing paradigm have been integrated with this approach. 
Edge computing with federated learning is one such application that has proven to be very successful. 
Edge devices have been used for multiple application with FL methods despite of its limited computational resources \cite{brecko2022federated}.
Training with different data space in FL with edge devices can achieve high accuracy by using horizontal and vertical federated learning.
Also, in hierarchical federated learning there can be multiple edge servers to perform partial model aggregation \cite{liu2020client}.

Edge computing with FL has vast application, such as in industry, healthcare, finance, transport etc \cite{brecko2022federated}.
Even for AI integrated mobile application, IoT systems like autonomous driving edge computing is being integrated with FL to deal with large amount of data and privacy issues.
Smart city sensing is another emerging paradigm where FL has great potential.
The valuable data can be used for developing0 smart AI application which can be beneficial in smart city infrastructure. Federated learning's ability to handle large scale data and security issues also provides significant advantages. Various approaches have been explored to enhance security, such as using blockchain with federated learning schemes for maintaining privacy \cite{jiang2020federated}. 

Although federated learning is a distributed machine learning algorithm, it maintains a single central server.
A challenge with this approach is that any failure in the central server can cause 
a breakdown of the whole architecture.
This issue has been mentioned in several researches but it is a open challenge to be solved\cite{baresi2022open}.
Our proposed approach can act as a fault tolerance in such situations.
By having multiple global servers, we can assure robust communication between client device and server device.
Multiple global model provides an option for the local devices to choose from based on the availability of the server.

Federated learning is a promising approach that is being used in various fields such as industries, IoT, and smart city infrastructure.
Its distributed architecture can handle large scale devices and preserves the privacy and security of data in these applications.
However, the previous researches rely on only one single central server and to reduce fault tolerance our proposed solution would handle multiple central server.

\section{Background} \label{preli}
\begin{table*}[t]
\caption{Comparison of aggregation model}   
\label{tab:ag_mod}
\begin{tabular}
{p{0.095\linewidth}p{0.13\linewidth}p{0.13\linewidth}p{0.13\linewidth}p{0.38\linewidth}}
\hline\hline
\textbf{   } & \textbf{Mechanism} & \textbf{Advantages} & \textbf{Limitations} & \textbf{Mathematical Expression} \\
\hline
\textbf{FedAvg\cite{mcmahan2017communication}\cite{konevcny2016federated}} & Averaging of local model updates from all participating nodes & Simple and easy to implement & Convergence may be slow due to communication bottleneck & \(w^{k+1} = \sum_{i{\in}C} \frac{n_i}{N}w_i^k\) \\ 
\hline
\textbf{FedAvgM\cite{sattler2019robust}\cite{chen2018federated}} & Averaging of local model updates with momentum & Faster convergence than FedAvg & Requires additional hyper-parameters to be tuned & \(v^k =\frac{\eta}{C} \sum_{i{\in}C}{\nabla}L_i(w^{k-1}) \newline w^k = w^{k-1}+v^k\) \\
\hline
\textbf{FedAdaGrad\cite{yu2020heterogeneous}} & Averaging of local model updates with adaptive learning rate & Fast convergence and can handle non-i.i.d data & Requires additional hyper-parameters to be tuned & \(\gamma_i^k = \gamma_i^{k-1}+({\nabla}L_i(w^{k-1}))^2 \newline w^k = w^{k-1}-\frac{\eta}{C}\sum_{i{\in}C}\frac{1}{\sqrt{\gamma_i^k}}{\nabla}L_i(w^{k-1})\) \\
\hline
\textbf{FedYogi\cite{li2022federated}} & Averaging of local model updates with adaptive learning rate & Handles non i.i.d data and noisy gradients well & Requires additional hyper-parameters to be tuned & \(w^{k=1}=\frac{\sum_{i{\in}C}n_iw_i^{k+1}}{\sum_{i{\in}C}n_i} \newline \theta^{k+1}=\theta^k-\frac{\eta}{\sqrt{v_t}+\epsilon}\sum_{i{\eta}C}\frac{n_i}{N}g_i^{k+1}\) \\
\hline
\textbf{FedAdam\cite{reddi2020adaptive}} & Averaging of local model updates with adaptive learning rate & Fast convergence, handles non-i.i.d data, and noisy gradients & Requires additional hyper-parameters to be tuned  & \(\theta^{k+1}=\theta^k-\eta(\frac{\sum_{i{\in}C}n_i}{N}^{-1})\sum_{i{\in}C}\frac{n_i}{N}(\frac{g_i^{k+1}}{\sqrt{\frac{1}{n_i}\sum_{j{\in}C}({g_j^k})^2+\epsilon}})\) \\
\hline

\end{tabular}

\end{table*}
In traditional federated learning approach, multiple devices create local models that are sent to the central server and aggregated into one global model.
The general theory can be stated as follows: 

\begin{equation} \label{eq_fl}
F(w) = \sum_{i=1}^{N}f_i(w)
\end{equation}

where, \(w = w_1, w_2, w_3, ... , w_N\).
\(f(w)\) represents the objective function, which is the sum of the local loss functions \(f_i(w)\) across all \(N\) clients.
\(w\) represents the global model parameters that need to be optimized.
\(w = w_1, w_2, w_3, ... , w_N\) represent the local model parameters on each client.
The optimization problem aims to minimize the aggregate loss or error across all clients while updating the global model iteratively based on local updates.

Federated learning uses different aggregation methods to combine the parameters from the local devices. 
The most used aggregation method is Federated Average (FedAvg), where the global model is created by using weighted averaged method \cite{sannara2021federated}.
The parameters from the local models would have more influence depending on the amount of data they been trained on.
Alternative aggregation methods have been proposed for federated learning. We provide a comparison of these aggregation methods in Table \ref{tab:ag_mod}.

\section{Proposed Methodology} \label{prop_meth}
\begin{figure*}[h!]
    \centering
    \includegraphics[scale=0.52]{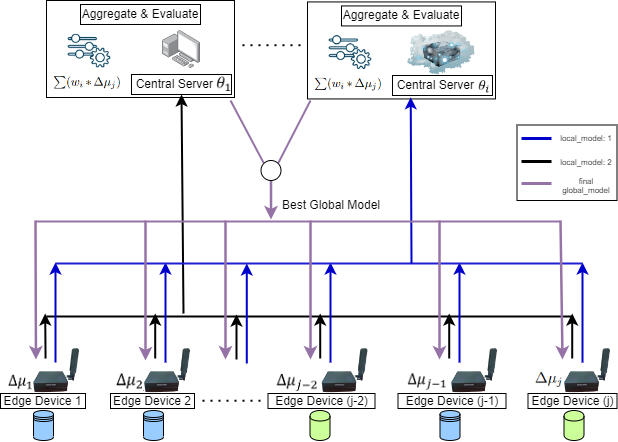}
    \caption{Multi-Global Server Architecture for Federated Learning}
    \label{fig:arch}
\end{figure*}

In our proposed approach, the data across multiple clients or devices are different, and the concept of multiple global server setting is introduced to each client. This approach aims to aggregate and update global model parameters based on the heterogeneous data from different clients.
Fig. \ref{fig:arch} shows the overall architecture of the proposed method.

\begin{table*}[bp]
    \centering
    \caption{EV recharge events data}
    \begin{tabular}
    {p{0.02\linewidth}p{0.20\linewidth}p{0.13\linewidth}p{0.06\linewidth}p{0.38\linewidth}}
\hline\hline
    \textbf{\#} & \textbf{Column Name} & \textbf{Non-Null Count} & \textbf{Dtype} & \textbf{Examples} \\ \hline
    0 & Connection ID & 11273 non-null & string & ab453504-a3b9-4c99-b238-e7ba374aa2f8 \\ \hline
    1 & Recharge Start Time (local) & 11273 non-null & datetime & 01/01/2020 11:05\\ \hline
    2 & Recharge End Time (local) & 11273 non-null & datetime & 01/01/2020 11:12\\ \hline
    3 & Account Name & 4423 non-null & string & ="" 	\\ \hline
    4 & Card Identifier & 11273 non-null & string & MA01091995436141\\ \hline
    5 & Recharge duration (hours:minutes) & 11273 non-null & datetime & 0:06\\ \hline
    6 & Connector used & 11273 non-null & string & J1772\\ \hline
    7 & Start State of charge (\%) & 11273 non-null & int & 4\\ \hline
    8 & End State of charge (\%) & 11273 non-null & int & 80\\ \hline
    9 & End reason & 11273 non-null & string & The charging cable was disconnected and put back in the station.\\ \hline
    10 & Total Amount & 11273 non-null & float64 & 0\\ \hline
    11 & Currency & 0 non-null & float64 & \\ \hline
    12 & Total kWh & 11273 non-null & float64 & 0.77\\ \hline
    13 & Station & 11273 non-null & string & NBA-10008\\ \hline
\end{tabular}


    \label{tab:data_sam}
\end{table*}
Mathematically, the update equation for multiple global server setting in federated learning can be represented as:
\begin{equation} \label{eq_mfl}
\theta_i(t+1) = \theta_i(t) + \eta * \sum(w_i * \Delta\mu_j)
\end{equation}
where, \(\theta_i(t+1)\) represents the updated global model parameters at time step \(t+1\).
\(\theta_i(t)\) represents the current global model parameters at time step t.
\(\eta\) is the learning rate, which controls the step size for parameter updates.
\(\sum(w_i * \Delta\mu_j)\) represents the weighted sum of the local model parameter updates from different clients.
\(w_i\) represents the weight assigned to the local model update \(\Delta\mu_j\) from client j. The weights can be determined based on factors such as the size of client data, the reliability of the client, or other metrics.
\(\Delta\mu_j\) represents the local model parameter update computed by client j, which captures the gradient or parameter adjustment specific to its local data.

By incorporating the weighted sum of local model updates from multiple clients, the global model can be iteratively improved while accounting for the heterogeneity in the data distribution across different clients.
The learning rate \(\eta\) determines the influence of the local updates on the global model, allowing for adaptation to the varying data characteristics.
\begin{table*}[t]
    \centering
    \caption{EV station details}
    \begin{tabular}
    {p{0.02\linewidth}p{0.20\linewidth}p{0.40\linewidth}p{0.20\linewidth}} \hline\hline
    \textbf{\#} & \textbf{Column Name} & \textbf{Description} & \textbf{Examples} \\ \hline
    0 & Location \# & Number indicating station & 1 \\ \hline
    1 & Business Location & Short description of business location(e.g city or business name) & Fredericton City Hall\\ \hline
    2 & Civic Address & The combination of the building number, street name and jurisdiction. & 397 Queen St, Fredericton\\ \hline
    3 & Station Name (separated by Type) & Unique identifier for a charging station & NBA-017 (L2 Station Name)\\ \hline
    4 & Rate (depending on Type) & Rate per hour for a charging station & 1.50/hr\\ \hline
    5 & GPS Coordinate & Latitude and longitude coordinates of the charging station & 45.964141, -66.643130\\ \hline
\end{tabular}

    \label{tab:data_sam1}
\end{table*}
\section{Experiment} \label{exp}
In this section, we implement our proposed method using a tabular dataset and a federated learning setup which we describe in detail.
\subsection{Dataset}
In our study, we used an electric vehicle charging event dataset. 
The dataset was provided by New Brunswick Power Consumption (NB Power), which contains the event histories from April 2019 to June 2022 \cite{richard2020discovering,richard2021spatial}.
We prepossessed the data by removing errors, inconsistency, redundancy, and duplicate or missing data.
Any columns with zero values were removed, and irrelevant symbols were deleted.
Also, multiple data files were combined to achieve consolidation and feature increment.
Table \ref{tab:data_sam} represents the raw recharge events of EV data.

We further labelled the energy consumption depending on the day of the week and the period of the day and added attributes such as station locations and level of charging stations as shown in Table \ref{tab:data_sam1}.
Afterwards, we did one-hot encoding to transform all the categorical values into numeric values for better regression.
In order to obtain an unbiased model, we removed any personal information of the clients from the dataset.
The dataset consists of various locations, which we used to distribute the dataset according to the region in edge clients.

\begin{figure}[h!]
    \centering
    \includegraphics[scale=0.22]{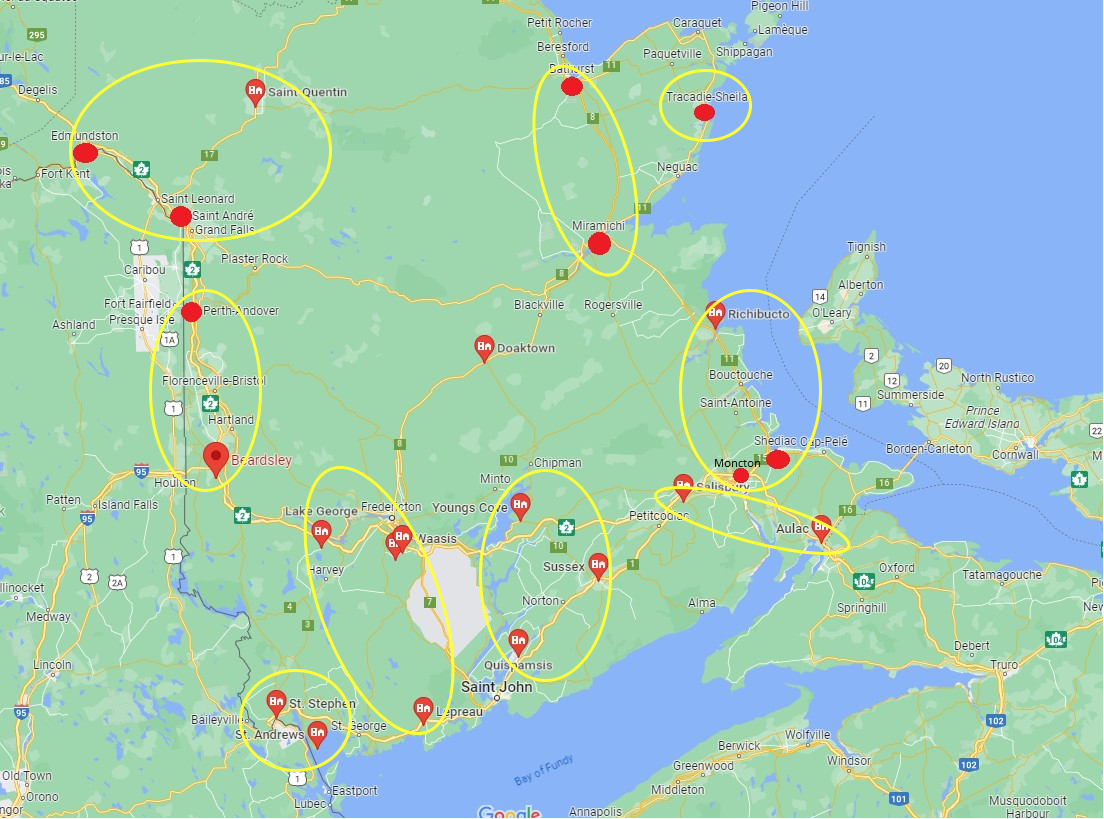}
    \caption{Map of EV stations}
    \label{fig:map}
\end{figure}

In Fig. \ref{fig:map}, the locations are shown and the groups have been created according to their GPS coordinates.

\subsection{Implementation}
Our setup consists of two global servers. For the first server, we used an Ubuntu cloud server, and Intel(R) Core(TM) i7-4790 CPU served as another global server. For simulating edge clients, we used a reComputer-Edge AI Device as a local device.
We split the EV dataset into nine regions across the simulated edge clients and trained the local model using the regression technique.
We divided and distributed the dataset depending on the station location so that each edge client had region-specific data, shown in Fig \ref{fig:map}.
After distributing the data in tabular form, we trained the local model in the edge client.
For local model training on edge clients, we used TabNet \cite{arik2021tabnet}.
TabNet is specifically designed for tabular datasets; it is composed of an attentive transformer, a feature transformer, and feature masking at each decision step.

Following the local training, the parameters from the local models are then sent to the global servers.
Each global server may or may not be connected with all the local devices, but each one has multiple edge client devices.
The parameters would be different as the distribution may vary in the client devices.
Getting all the local parameters, the central servers use the FedAvg algorithm to aggregate the parameters and produce a global model.
As there would be multiple global models, we evaluated and compared the loss functions of each of the global models with one another.

For the error tolerance, we implemented a simple algorithm containing a 'try-except' block to switch the global server.
Let us consider that there are n number of servers, and the number of servers and server addresses will be the input.
The details are shown in the Algorithm \ref{alg:com}, where the system tries to establish a connection between the client device and the available server to send the local parameters.

\begin{algorithm}[]
\caption{Connect to Global Server}\label{alg:com}
\textbf{Input:} \text{List of server addresses S[1..n], Number of servers n} \hspace*{\fill} \\
\textbf{Output:} \text{Connection Status}\hspace*{\fill}

\algblock[TryCatchFinally]{try}{endtry}
\algcblock[TryCatchFinally]{TryCatchFinally}{finally}{endtry}
\algcblockdefx[TryCatchFinally]{TryCatchFinally}{catch}{endtry}
	[1]{\textbf{catch} #1}
	{\textbf{end try}}

\begin{algorithmic}[1]
    \State{Initialize connectionEstablished = False}
    \State{Initialize selectedServer = None}
    
    \For{i from 1 to n}
        \try
            \State{Connect to S[i]}
            \State{Set connectionEstablished to True}
            \State{Set selectedServer to S[i]}
            \State{Break}
        \catch{connectionFailed}
            \State{Continue to the next server}
        \endtry
        \If{connectionEstablished is True}
            \State{Send local parameters to selectedServer}
        \Else
            \State{Print "Failed to connect to all servers."}
        \EndIf
    \EndFor
    
\end{algorithmic}
\end{algorithm}
\section{Preliminary Results} \label{result}
\begin{figure}[h!]
    \centering
    \includegraphics[scale=0.45]{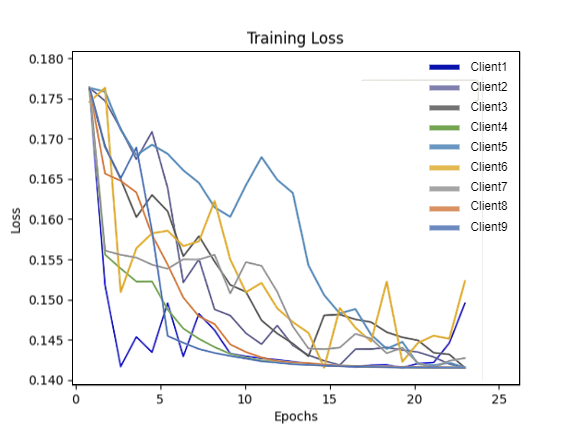}
    \caption{Training Loss: Edge Clients}
    \label{fig:loss1}
\end{figure}

The preliminary results of our experiment are provided herewith. 
We ran 25 epochs on each client and observed their training loss while producing the local parameters. The plots of training loss are shown in Fig \ref{fig:loss1}.
Each of the edge clients began with a high training loss, but eventually, the loss dropped to 0.1\%.
For instance, Client5 had an increase in the loss after 10 epochs but at the 25th epoch, the loss decreased significantly.
On the other hand, Client8 had a more smooth loss reduction while training.
At the end of the training, each of the clients produced local parameters. 

\begin{figure}[h!]
    \centering
    \includegraphics[scale=0.5]{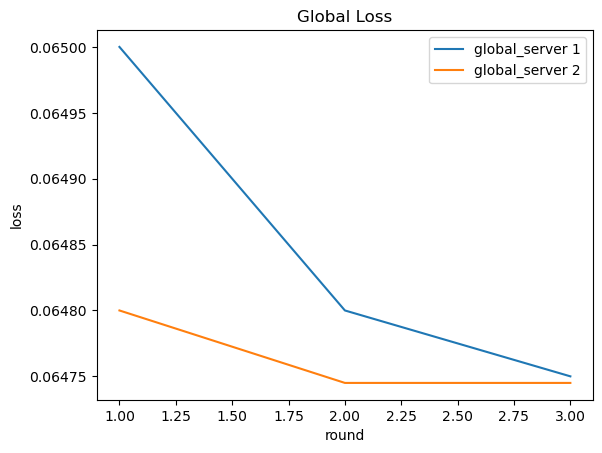}
    \caption{Global Loss Comparison}
    \label{fig:glob_L}
\end{figure}

The local parameters were then sent to both the global servers to produce global models.
In Fig \ref{fig:glob_L}, we show the loss of two global servers.
We refer to the first server (Ubuntu cloud server) as \textit{global\_server1} and the other server (Intel(R) Core(TM) i7-4790 CPU) as \textit{global\_server2}.
Each server conducted a three-round aggregation with the client or local model parameters.
After each round of aggregation, the global model updated and evaluated its loss.
The difference in losses in Fig \ref{fig:glob_L} was observed to be considerably small, and we can see that \textit{global\_server2} performs slightly better in aggregating the models. Even though the difference in loss between the global models was initially large, they both converge to a similar loss value at the third round. 

\section{Conclusion and Future Work} \label{conc}
There are numerous industrial uses of single global server federated learning approach with image datasets, but using tabular data is comparatively rare.
Our proposed system leverages multi-global server FL architecture with a tabular dataset.

Multiple global servers in federated learning ensure robustness to development challenges and error-tolerance.
In this paper, we used an electric vehicle dataset in a federated learning architecture with multiple global servers and predicted the energy consumption. We compared our multiple-global server approach against a single-global server approach and observed that our method had less than 1\% increase in performance. Also, the training efficiency did not vary as expected in terms of computational speed and loss. However, our results show that communication challenges can be mitigated by implementing multiple global servers without a significant loss in performance.

For our future work, we will introduce a decision device to dynamically handle client-server communication and facilitate the federated learning process.
A holistic evaluation encompassing a broader spectrum of metrics, including energy consumption, model robustness, and scalability, is another research direction, which will help understand the overall implications and trade-offs associated with employing multiple global servers in federated learning.

\section*{Acknowledgment}
This work is supported by the NBIF Talent Recruitment Fund (TRF2003-001) and partially supported by The Harrison McCain Young Scholars Awards. We would like to thank NB Power for providing us with the data. We also would like to thank Rene Richard for our fruitful conversation.






\bibliographystyle{ieeetr}
\bibliography{sections/bibliography.bib}



\end{document}